\documentclass{article}

\usepackage{arxiv}

\usepackage[utf8]{inputenc} 
\usepackage[T1]{fontenc}    
\usepackage{hyperref}       
\usepackage{url}            
\usepackage{booktabs}       
\usepackage{amsmath,amssymb,amsfonts}       
\usepackage{nicefrac}       
\usepackage{microtype}      
\usepackage{lipsum}
\usepackage{graphicx}
\usepackage{algorithmic}
\usepackage{algorithm}

\title{Autonomous Deep Learning: Continual Learning Approach for Dynamic Environments \thanks{This paper has been published in Proceedings of the 2019 SIAM International Conference on Data Mining. The source code is available in \url{https://www.researchgate.net/publication/335757711_ADL_Code_mFile}.}}

\author{
  Andri Ashfahani\thanks{Equal contribution.} \\
  School of Computer Science and Engineering\\
  Nanyang Technological University\\
  Singapore, 639798 \\
  \texttt{andriash001@e.ntu.edu.sg} \\
   \And
 Mahardhika Pratama $^\dagger$ \\
  School of Computer Science and Engineering\\
  Nanyang Technological University\\
  Singapore, 639798 \\
  \texttt{mpratama@ntu.edu.sg} \\
}

\begin{document}
\maketitle

\begin{abstract}
The feasibility of deep neural networks (DNNs) to address data stream problems still requires intensive study because of the static and offline nature of conventional deep learning approaches. A deep continual learning algorithm, namely autonomous deep learning (ADL), is proposed in this paper. Unlike traditional deep learning methods, ADL features a flexible structure where its network structure can be constructed from scratch with the absence of initial network structure via the self-constructing network structure. ADL specifically addresses catastrophic forgetting by having a different-depth structure which is capable of achieving a trade-off between plasticity and stability. Network significance (NS) formula is proposed to drive the hidden nodes growing and pruning mechanism. Drift detection scenario (DDS) is put forward to signal distributional changes in data streams which induce the creation of a new hidden layer. Maximum information compression index (MICI) method plays an important role as a complexity reduction module eliminating redundant layers. The efficacy of ADL is numerically validated under the prequential test-then-train procedure in lifelong environments using nine popular data stream problems. The numerical results demonstrate that ADL consistently outperforms recent continual learning methods while characterizing the automatic construction of network structures.
\end{abstract}

\section{Background and Motivation}
State-of-the-art theoretical studies show that the increase of depth of neural networks increases the representational and  generalization power of neural networks (NNs) \cite{NIPS2012_4824,IMM2012}. Nevertheless, the problem of data stream remains an uncharted territory of conventional deep neural networks (DNNs). Unlike conventional data stream methods built upon a shallow network structure \cite{DBLP:journals/corr/abs-1805-07715,pensembleplus,pENsemble}, DNNs potentially offers significant improvement in accuracy and aptitude to handle unstructured data streams. Direct application of conventional DNNs for data stream analytic is often impossible because of their considerable computational and memory demand making them impossible for deployment under limited computational resources \cite{GamaDataStream}. Ideally, the data streams should be handled in a sample-wise manner without any retraining phase to prevent the catastrophic forgetting problem in  addition to scale up with the nature of continual environments \cite{GamaDataStream,DeepExpandable}. Another challenge comes from the fixed and static structure of traditional DNNs \cite{Gamaconceptdrift}. In other words, the network capacity has to be estimated before process runs. This trait does not mirror the dynamic and evolving characteristics of data streams. 

The use of flexible structure with the growing and pruning mechanism has picked up research attention in DNN literature \cite{pENsemble,pensembleplus,DEVFNN} where the key idea is to evolve the DNN's structure on demand. Incremental learning of denoising autoencoder (DAE) realizes the structural learning mechanism via the network's loss and the hidden unit merging mechanism \cite{Zhou_incrementallearning}. The underlying drawback of this approach is located in the over-dependence on problem-dependent predefined thresholds in growing and merging hidden units. The elastic consolidation weight (ECW) \cite{ECW} and the hedge backpropagation (HBP) \cite{OnlineDeepLearning} are proposed to train DNN in the online situation where the ECW method addresses the catastrophic forgetting problem by  preventing the output weights of new task to be deviated too far from the old one, while the HBP realizes a direct connection of hidden layer to output layer which enables representation of different concepts in each layer. However, these approaches call for network initialization step and operates under a fixed capacity. 

The progressive neural networks (PNN) \cite{DBLP:journals/corr/RusuRDSKKPH16}, the dynamically expandable networks (DEN) \cite{DeepExpandable} and incremental learning of DAE (DEVDAN) \cite{devdan} are proposed to address limited network capacity and catastrophic forgetting problems. PNN creates a new network structure for every new task, DEN grows hidden nodes whenever the loss criteria are not satisfied, while DEVDAN is capable of growing and pruning the hidden units based on the estimation of network significance (NS). Nevertheless, the three approaches utilize a fixed-depth structure \cite{Gamaconceptdrift,NIPS2012_4824}. It is understood from \cite{powerofdepth} that addition of network depth leads to more significant improvement of generalization power than addition of hidden unit because it boosts the network capacity more substantially. To the best of our knowledge, the three approaches have not been tested under the prequential test-then-train scenario which reflects a situation where data stream arrives without label \cite{GamaDataStream}.

\section{Problem Formulation}
\label{sec:headings}
Continual learning of evolving data streams is defined as learning approach of continuously generated data batches $B \in[B_1,\dots,B_k,\dots,B_K]$ where the number of data batches $K$ and the type of data distributions are unknown before the process runs. $B_k$ can be either a single data point $B_k = X\in\Re^{n}$ or a particular size of data batch $B_k = [X_1,\dots,X_t,\dots,X_T]\in\Re^{T\times n}$, where $n$ and $T$ denote the dimension of the input space and the number of data points in a batch, respectively. Note that the batch size often varies across different time stamps. In the data stream problems, data points come into picture with the absence of true class label \cite{GamaDataStream}. The execution of labelling process is subject to the access of the ground truth or expert knowledge. In other words, a delay is expected while revealing the true class labels $C\in\ \Re^{T}$. The $0-1$ encoding scheme can applied to obtain multi-output target matrix $C\in\ \Re^{T\times m}$ where $m$ is the number of target. This issue limits the feasibility of cross-validation or direct train-test partition methods as an evaluation protocol because those methods assume that the overall data batches are fully observable and risks on loss of data temporal order \cite{GamaDataStream,liu2018accumulating}.

The data streams require DNN to handle $B_k$ which may be originated from different data distributions, also known as the concept drift. Specifically, there may exist a change of joint-class posterior probability $P(C_t,X_t)\neq P(C_{t-1},X_{t-1})$. The concept drift is commonly classified into two types: real and covariate \cite{Gamaconceptdrift}. The real drift usually is more severe than the covariate drift because the input variations lead to the shift of decision boundary which decreases the classification performance. In addition, this leads to a model created by previous concept $B_{k-1}$ being outdated. This characteristic shares some relevance with the multi-task learning problem where each data batch $B_k$ is of different tasks. Nevertheless, DNN differs from the multi-task approaches in which all data batches are to be processed by a single model rather than rely on task-specific classifiers. Another problem of data streams exists in achieving trade-off between plasticity and stability which increases the risk of suffering from catastrophic forgetting \cite{kirkpatrick2016overcoming}. These demands call for an online DNN model which is capable to incrementally construct its network structure from scratch in respect to data streams distribution. In addition, a mechanism to flexibly reuse and retain the old knowledge, or to learn the new one should be embedded to prevent catastrophic forgetting.

\section{Our Approach}
A fully elastic deep neural network (DNN), namely Autonomous Deep Learning (ADL), is proposed in this paper. ADL features an open structure where  not only its hidden nodes can be self-organized but also the hidden layers can be constructed under the lifelong learning paradigm. These mechanisms enable ADL to perform dynamic resource allocation which tracks the dynamic variation of data streams \cite{DBLP:journals/corr/RusuRDSKKPH16,DeepExpandable}. The adaptation of network width is governed by network significance (NS) method which governs creation of new hidden units and pruning of inconsequential hidden units. The adaptation of network depth is driven by drift detection scenario (DDS) where a new hidden layer is added if a drift is identified. Every hidden layer embraces different concepts played in different time windows of data streams \cite{Bengio_2013}. The complexity reduction mechanism in the hidden layer level is implemented through the hidden layer merging procedure which quantifies mutual information of hidden layers and coalesces those suffering from high mutual information \cite{DEVFNN}. A new DNN structure is introduced where it puts into perspective the different-depth concept. That is, every layer is connected to a softmax layer which produces a local output. The global output is obtained from aggregation of each local output using the dynamic voting scenario. The generalization power of ADL is evaluated under \textbf{the prequential test-then-train protocol with only a single epoch} where the data are first use to test before exploited to update the model.The major contributions are elaborated as follows:
\paragraph{Different-depth network structure.} Unlike traditional DNN structure, where the final output relies on the last hidden layer, ADL puts forward the different-depth structure where there exists a direct connection of each layer to the output layer by inserting a softmax layer in each hidden layer to produce a layer-specific output. The dynamic voting scheme is integrated to deliver the final classification decision where every layer is assigned with a voting weight adapted with different intensities in respect to layer's relevance. This approach is capable of overcoming the catastrophic forgetting problem because a network structure is constructed as a complete summary of data distributions \cite{DBLP:journals/corr/RusuRDSKKPH16}. Moreover, the dynamic voting weight mechanism is designed with dynamic decaying rates in respect to the prequential error which enables the strongest layer to dominate the voting process.
\paragraph{Network width adaptation.} ADL features elastic network width which supports automatic generation of new hidden nodes and pruning of inconsequential nodes. This mechanism is controlled by the NS method \cite{devdan} which estimates the network generalization power in terms of bias and variance. A new hidden node is added in the case of underfitting (high bias) while the pruning mechanism is activated in the case of overfitting (high variance). Another salient feature of NS is not dependent on the user-defined parameters which enables the plug-and-play operation. It uses an adaptive threshold which dynamically adapts to the bias and variance estimation. This work offers an extension of \cite{devdan} for a deep network structure. 
\paragraph{Network depth adaptation.} The drift detection scenario (DDS) is employed to self-organize the depth of network structure where the depth of network structure increases if a drift is signalled. This idea is supported by the fact that addition of hidden layer induces more active regions than addition of hidden units, thereby being able to rectify the high bias situation due to drift effectively \cite{linearregion}. Note that active region here refers to the amount of unique representation carried by a hidden layer. In other words, DDS guides ADL to arrive with the hidden layers carrying the different concepts of data streams. Furthermore, the DDS method detects the real drift - variation of input space causing variation of output space via the evaluation of accuracy matrix based on the Hoeffding's bounds method \cite{frias2015online}. ADL also implements the complexity reduction scenario shrinking the depth of network structure. This scenario is achieved by the analysis of mutual information across hidden layers. A hidden layer sharing high correlation is discarded. This concept follows \cite{pensembleplus} but here this concept is played under the context of DNN.
\paragraph{Solution of catastrophic forgetting.} The key property of ADL in addressing the catastrophic forgetting problem lies in the different-depth architecture which allows to accommodate new knowledge while revisiting old knowledge with ease \cite{DBLP:journals/corr/RusuRDSKKPH16}. Moreover, the final output is produced by the dynamic voting scheme which enables to flexibly give more emphasis either to the old knowledge or to the new ones. This is evident because each layer is assigned with unique voting weights which increases and decreases with different rates. Moreover, the parameter tuning process is localized to the most relevant concept, namely the winning layer while freezing other layers to assure stable old concepts - old concept is not perturbed.

\section{ADL Learning Policy}
\begin{figure}[t!]
	\begin{centering}
	\includegraphics[scale=0.6]{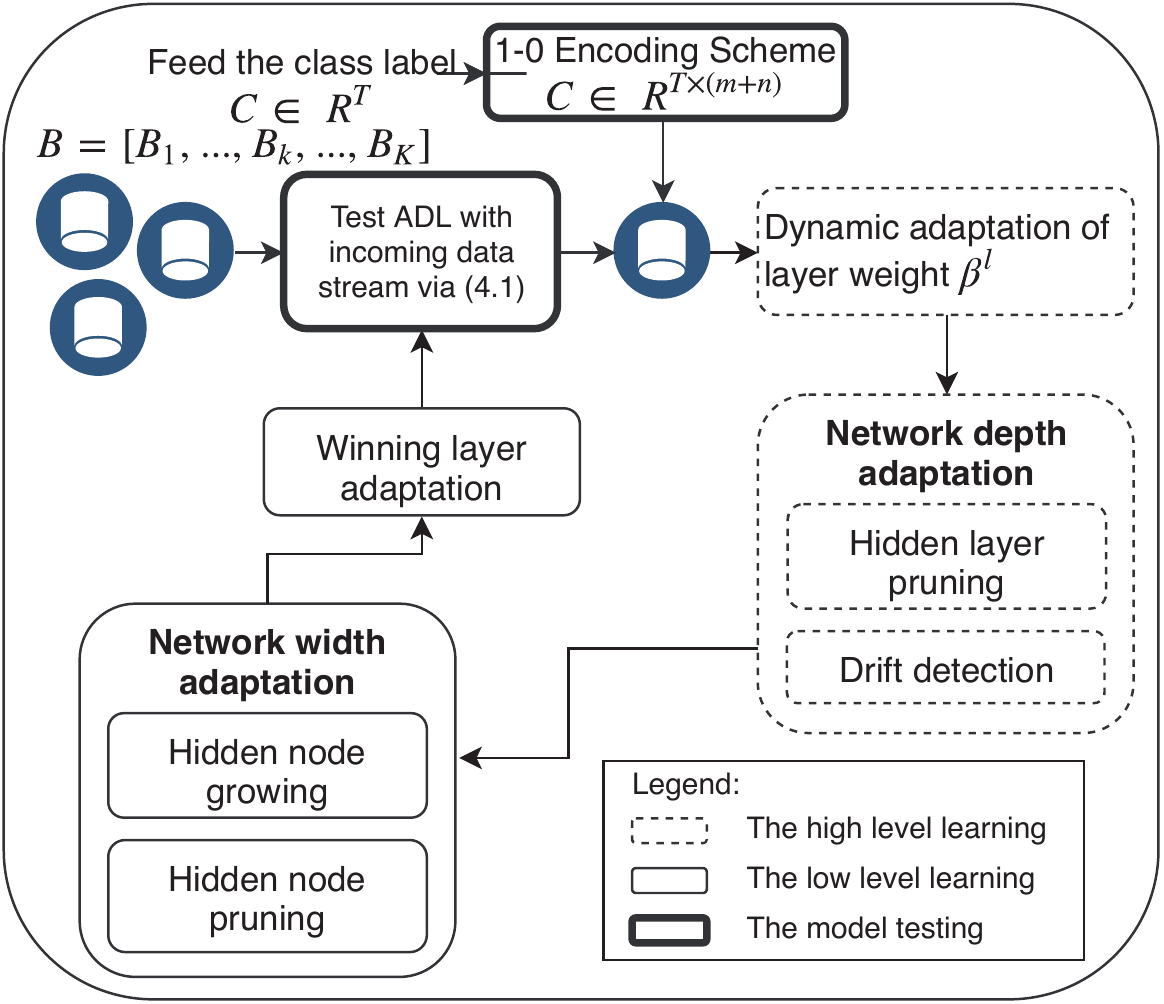}
	\par\end{centering}
	\caption{The learning policy of ADL}
	\label{fig:adllearning}
\end{figure}

This section explains the network structure of ADL and its learning policy, which is depicted in Figure \ref{fig:adllearning}.

\subsection{Network Structure and Working Principle}
The ADL is constructed by the multilayer perceptron (MLP). The first layer defines the input feature, while the intermediate layers consists of multiple linear transformations  interspersed by sigmoid function. The hidden layers and the hidden nodes of ADL can be automatically constructed which are controlled by DDS and NS formula, respectively. ADL characterizes the different-depth structure formalized as follows:
\begin{eqnarray}\label{equation:C}
    \hat{C} = \underset{o=1,\dots,m}{\max}\hat{Y_o};\,\,\hat{Y}=\sum_{l=1}^{L}\beta^{(l)}.y^{(l)}\\
    y^{(l)} = s.max(Ws^{(l)}h^{(l)}+bs^{(l)}),\,\,\forall l = 1,\dots, L\nonumber\\
    h^{(l)} =\sigma(W^{(l)}h^{(l-1)}+b^{(l)}),\,\,h^{(0)}=X\nonumber
\end{eqnarray}
From (\ref{equation:C}), it is observed that every hidden layer $h^{(l)}$ has a connection to a unique classifier producing the multiclass probability $y^{(l)}$, where $L$ is the number of hidden layers. The network parameters of $l$-th hidden layer are denoted as $\theta^{(l)}$ that is, $W^{(l)} \in\Re^{R_l \times d}$, $b^{(l)}\in\Re^{R_l}$, $Ws^{(l)} \in\Re^{m \times R_l}$, $bs^{(l)}\in\Re^{m}$, where $R_l$ and $d$ is the number of hidden nodes and the number of input in the $l$-th hidden layer, respectively. It is worth noting that the dimension of those matrices is changing according to the evolution of hidden nodes. The hidden layer is assigned with a voting weight $\beta^{(l)}$ which is dynamically adjusted by a dynamic penalty and reward factor $p^{(l)}$. The voting weights are normalized, $\sum_{l=1}^{L}\beta^{(l)}=1$, to ensure the partition of unity. Finally, the predicted label is the class label embracing the highest $\hat{Y}_o,\forall o=1,\dots,m$, obtained by combining the weighted hidden layer output, as per in (\ref{equation:C}).

ADL starts its learning process from scratch without initial structure. ADL here is simulated under the prequential test-then-train procedure where data stream is first used for the testing process followed by the training process. This scenario realizes the fact that data stream come unlabelled. ADL consists of two learning stages: the high level learning and the low level learning. The former one concerns on  the evolution of hidden layer while the later stage focuses on the network parameters and the number of hidden nodes of the winning layer $l_w$ using SGD and NS formula, respectively, in a single-pass learning fashion. The winning layer is a hidden layer embracing the highest $\beta$. The voting weight is deemed as an appropriate indicator of the hidden layer performance since it is adjusted using dynamic factor. Generally, the low level learning enables ADL to learn new knowledge while retaining the old ones. Moreover, it helps ADL to handle the virtual-drift, that is a distributional change of the input space \cite{Gamaconceptdrift}. After executing the low level learning, the generalization performance of ADL is evaluated using the labelled data batch $B_k=[X_k,C_k]\in\Re^{T\times(n+m)}$.

The evaluation results are then exploited in the high level learning process which consists of three mechanisms. The first one is dynamic voting weight adaptation. Every $y^{(l)}$ will be penalized if it makes an incorrect prediction and, conversely, it will be rewarded if it makes a correct prediction using dynamic penalty and reward factor $p^{(l)}$. Secondly, hidden layer pruning scenario is carried out to discard the redundant hidden layer. It is defined as the $l_p$-th hidden layer, $y^{(l_p)}$, which is highly correlated to others yet it has low performance. The MICI method \cite{pensembleplus} is employed to explore the mutual correlation of hidden layer, $\gamma(y^{(i)},y^{(j)}),\forall i,j=1,\dots,L,\,i\neq j$. The first two mechanisms enable ADL to ignore the less useful representations and to emphasize the useful ones while obtaining the predicted output $\hat{C}$. Lastly, network depth adaptation is conducted by executing DDS. This method monitors the statistics of accuracy matrix and categorizes the behaviour into three stages, i.e., stable, warning, and drift. A new hidden layer is constructed when a drift is confirmed. The voting weight of newly created layer $\beta^{(L)}$ and its decreasing factor $p^{(L)}$ are set to 1, while the network parameters $\theta^{(L)}$ are initialized via the low level learning phase using the current data batch. The last adjustment aims to increase the generalization and representational power of ADL. Figure \ref{fig:evolution} exemplifies the overall incremental learning process of ADL where $[n,m]=[3,2]$.

\subsection{Network Width Adaptation}
This policy is carried out in the low level learning process which consists of two mechanisms as follows.
\paragraph{Hidden node growing.} The hidden node growing mechanism is controlled by the NS formula which evaluates the generalization power of network structure formalized as the expectation of squared error under a normal distribution as per in (\ref{equation:ns}). This expression leads us to the bias-variance formula as per in (\ref{equation:biasvar}). 
\begin{eqnarray}
    NS=\int_{-\infty}^{\infty}(C-\hat{y}^{(l_w)})^{2}p(x)dx \label{equation:ns}\\
    NS=Var(\hat{y}^{(l_w)})+(Bias(\hat{y}^{(l_w)}))^2\label{equation:biasvar},\\
    NS=(E[(\hat{y}^{(l_w)})^{2}]-E[\hat{y}^{(l_w)}]^{2})+(E[\hat{y}^{(l_w)}]-C)^{2}\nonumber
\end{eqnarray}

The solution of (\ref{equation:ns}) requires to calculate $E[\hat{y}^{(l_y)}]$. Note that $\hat{y}^{(l_w)}$ is the deterministic function of $X$ that is, the input of ${h}^{(1)}$. Therefore, the key to solve the definite integral in (\ref{equation:ns}) is the solution of $E[{h}^{(1)}]=\int_{-\infty}^{\infty}\sigma(W^{(1)}X+b^{(1)})p(X)dX$. Suppose $X$ possesses normal distribution, the probability density function $p(X)$ is given as $\frac{1}{\sqrt{2\pi\sigma^2}}\exp(-\frac{(\widetilde X - \mu)^{2}}{2\sigma^{2}})$, where $\mu$ and $\sigma$ are the recursive mean and recursive standard deviation of data streams, $X$, which can be calculated easily. It is worth noting that a sigmoid function can be approximated by a probit function $\Phi(\xi X)$ where $\Phi(X)=\int_{-inf}^{X}\mathcal{N}(\theta|0,1)d\theta$ and $\xi=\frac{\pi}{8}$. The integral of probit function is another probit function \cite{Murphy_Machine_Learning}, it yields:
\begin{equation}\label{equation:probit}
    E[{h}^{(1)}]=\sigma(W^{(1)}\mu/(\sqrt{1+\pi\sigma^{2}/8})+b^{(1)})
\end{equation}
Next, $E[\hat{y}^{(l_w)}]$ can be obtained by generalizing (\ref{equation:probit}) via $l_w$ times-forward-chaining operation. It yields:
\begin{gather}
E[\hat{y}^{(l_w)}]= s.max(Ws^{(l_w)}E[h^{(l_w)}]+bs^{(l_w)})\label{equation:Ey} \\
E[h^{(l)}] =\sigma(W^{(l)}E[h^{(l-1)}]+b^{(l)}),\,\forall l = 2,\dots, l_w \label{equation:HS}
\end{gather}
After that, the bias can be calculated by substituting $E[\hat{y}^{(l_w)}]$ to the bias term, $(Bias(\hat{y}^{(l_w)}))^2 = (E[\hat{y}^{(l_w)}]-C)^2$. This approach is different from the loss function used in \cite{DeepExpandable}, because while approximating the generalization of DNN, the bias formula takes into account the influence of all past and future samples under the assumption of normal distribution. The high bias indicates the underfitting situation which can be circumvented by increasing the network capacity.

The hidden node growing condition is derived based on $k$-sigma rule concept adopted from the theory of statistical process control \cite{Gama2006}. However, instead of using the binomial distribution to calculate the mean and variance, ADL directly utilizes the bias itself ($Bias^2$) because the hidden node growing strategy evaluates the real-variable bias instead of the accuracy score. The high bias problem, triggering the construction of a hidden node in the $l_w$-th layer, is formulated as follows:
\begin{equation} \label{equation:grow}
    \mu_{bias}^{t}+\sigma_{bias}^{t} \geq \mu_{bias}^{min}+\pi\sigma_{bias}^{min}
\end{equation}
where $\pi=1.3exp(-(Bias(\hat{y}^{(l_w)}))^2)+0.7$ and governs the confidence degree of sigma rule. It is designed that $\pi$ is a function of $Bias^2$ and revolves around $[1,2]$. A high bias triggers $\pi$ to return a low confidence level - close to 1, realizing around $68.2 \%$ confidence degree. Conversely, a high confidence level - close to 2, equivalent to $95.2 \%$, is generated by $\pi$ when the bias is low. This provides a flexibility for the hidden nodes growing mechanism and eliminates the involvement of problem-specific threshold. $\mu_{bias}^{t},\sigma_{bias}^{t}$ are the recursive mean and standard deviation of $Bias^2$ up to $t$-th time instant, whereas $\mu_{bias}^{min},\sigma_{bias}^{min}$ denote the minimum value of $\mu_{bias},\sigma_{bias}$ up to $t$-th time instant but are reset whenever (\ref{equation:grow}) is satisfied. Equation (\ref{equation:grow}) signifies the existence of changing data distribution represented by the increase of network bias. The network bias should decrease or at least be stable when there is no drift in data streams. When (\ref{equation:grow}) is satisfied, a hidden node is added in the $l_w$-th hidden layer, $R_{l_w}=R_{l_w}+1$, and the new network parameters in the $R_{l_w}$-th hidden node $\theta^{(l_w)}_{R_{l_w}}$ are initialized using Xavier initialization.
\paragraph{Hidden node pruning.} It is derived from the same principle of the hidden node growing mechanism, yet it exploits $Var(\hat{y}^{(l_w)})$ instead of $(Bias(\hat{y}^{(l_w)}))^2$. A high variance, overfitting, should be handled by reducing the network complexity. Before calculating $Var(\hat{y}^{(l_w)})$, it is required to derive the expression of $E[(\hat{y}^{(l_w)})^{2}]$ and $E[\hat{y}^{(l_w)}]^{2}$. The second expression can be obtained easily by applying squared operation to $E[\hat{y}^{(l_w)}]$. It is worth noting that $(\hat{y}^{(l_w)})^2=\hat{y}^{(l_w)}*\hat{y}^{(l_w)}$ is the IID variable. Therefore, $E[(\hat{y}^{(l_w)})^{2}]$ can be obtained by first calculating $E[(h^{(1)})^2]=E[{h}^{(1)}]*E[{h}^{(1)}]$ and followed by forward-passing the result to $l_w$-th hidden layer. It is similar to the way of calculating (\ref{equation:Ey}), yet it takes $E[(h^{(1)})^2]$ as the initial input instead of $E[(h^{(1)})]$.

The hidden node pruning condition implements the same principal as the growing part where the statistical process control is adopted to identify the high variance problem, as per in (\ref{equation:prune}). Unlike the growing condition in (\ref{equation:grow}), the $\sigma_{var}^{min}$-part is multiplied by 2 to avoid direct-pruning-after-growing problem. It is worth mentioning that the addition of a hidden node leads to the increase of network variance yet progressively decrease as the next information arrives. $\chi$ is designed similar to $\pi$ yet it takes $Var(\hat{y}^{(l_w)})$ as the input instead of $(Bias(\hat{y}^{(l_w)}))^2$. Consequently, the $k$ sigma rule revolves in the range of $[2,4]$ providing around $95.4\%$ to $99.9\%$ confidence level.
\begin{equation} \label{equation:prune}
    \mu_{var}^{t}+\sigma_{var}^{t} \geq \mu_{var}^{min}+2\chi\sigma_{var}^{min}
\end{equation}

If (\ref{equation:prune}) is satisfied, the pruning scenario is undertaken to remove the weakest hidden node in the $l_w$-th hidden layer. The least significant hidden node can be observed by calculating (\ref{equation:HS}). That is, the importance of all hidden nodes in the $l_w$-th hidden layer. The pruning mechanism discarding the hidden node with the lowest $E[h^{(l_w)}]$ is formalized as follows: $Pruning \rightarrow \min_{i=1,...,R_{l_w}}E[h^{(l_w)}]_i$. Consequently, the number of hidden nodes decreases to $R_{l_w}=R_{l_w}-1$ as an effort to address the overfitting dilemma. Note that a small $E[h^{(l_w)}]_i$ value indicates that $i$-th hidden node plays a small role in producing the output $y^{(l_w)}$ and thus can be discarded without significance loss of accuracy. The concept of statistical contribution of hidden node can be categorized as performance estimation strategy of neural architecture search because it estimates the generalization power of the network on unseen data \cite{elsken2018neural}.

\subsection{Network Depth Adaptation}
ADL realizes the different-depth structure using the DDS as an effort to deal with the concept drift. It also utilizes the MICI method as the complexity reduction procedure. The following explain those mechanisms.

\paragraph{Hidden layer growing.} A new hidden layer is constructed if there is a concept change in the data streams.  The DDS signals a drift status by monitoring the accuracy matrix. The drift situation signifies that the network is underfitting as indicated by low statistic of accuracy matrix meaning that the ADL's performance deteriorates. In other words, the crafted knowledge alone cannot adequately describes the new data distribution. This dilemma can be solved by increasing the network capacity in two ways those are, hidden node growing or hidden layer expansion. The second option, however, augments the network capacity more significantly since the extension of depth increases the number of active regions \cite{linearregion} more than that expansion of network width. In addition, addition of network depth has been theoretically more meaningful than addition of neuron or hidden units \cite{powerofdepth}.

The accuracy matrix $F \in \Re^{T\times m}$ stores the generalization performance of the testing phase. It records 1 if the misclassification happens $\hat{C}_t \neq C_t$, whereas 0 is stored if ADL correctly classify an observation $\hat{C}_t = C_t$. The switching point is determined by evaluating two accuracy matrices, $F$ and $G\in \Re^{cut \times m}$, where $cut$ is the hypothetical switching point which can be found using the following condition:
\begin{equation}\label{equation:cuttingpoint}
    \hat{F}+\epsilon_F \leq \hat{G}+\epsilon_G
\end{equation}
where $[\hat{F}, \hat{G}]$ and $[\epsilon_F,\epsilon_G]$ denote the statistics and the Hoeffding's error bounds of $[F,G]$. The condition (\ref{equation:cuttingpoint}) spots a transition between two concepts where $\hat{G} > \hat{F}$. Note that $\hat{G}$ is expected to decrease or at least be constant in the stable phase. This strategy performs better while dealing with sudden drift, the most common type of drift, yet it is less sensitive to the gradual drift where change slowly appears because every sample is treated equally without any weights \cite{frias2015online}. The Hoeffding's error bounds are formulated as follows:
\begin{equation}\label{equation:errorbound}
    \epsilon_{F,G,H}=(b-a)\sqrt{\frac{size}{2(size\times cut)}\ln(\frac{1}{\alpha})}
\end{equation}
where $size$ denotes the size of accuracy matrices and $\alpha$ denotes the significance level of Hoeffding's bound. Note that $\alpha$ is statistically justifiable since it is associated to the confidence level $1-\alpha$. It is not classified as a problem-specific threshold because a high $\alpha$ provides a low confidence level whereas a low $\alpha$ returns a high one. The values $a,b$ indicate the minimum and the maximum entries of the accuracy matrices $F,G,H$.
\begin{gather}
    |\hat{H}-\hat{G}|>\epsilon_D\label{equation:drift}\\
    \epsilon_W \leq|\hat{H}-\hat{G}|<\epsilon_D\label{equation:warning}
\end{gather}

The condition (\ref{equation:cuttingpoint}) aims at finding the cutting point, $cut$, where the accuracy matrix $G$ is not in the decreasing trend. Once it is spotted, the accuracy matrix, $H \in \Re^{(T-cut)\times m}$, can be formed. This matrix is used as a reference whether the null hypothesis is valid or not. The null hypothesis evaluates the increase of statistics of accuracy matrix which verifies the drift condition. The drift status is signalled when the null hypothesis is rejected with the size of $\alpha_D$, as per in (\ref{equation:warning}). Conversely, the warning status is returned when the null hypothesis is rejected with the size $\alpha_W$, formalized in (\ref{equation:warning}), which aims to signal the gradual drift. The value of $\epsilon_W$ and $[\epsilon_D,\epsilon_F,\epsilon_G]$ can be calculated via (\ref{equation:errorbound}) using $\alpha_W$ and $\alpha_D$. If none of those conditions are satisfied, the stable condition is returned.

A drift condition (\ref{equation:drift}) displays the phase where the empirical mean of $G$ is lower than $F$ indicating the evidence that the classification performance degenerates. This case signals the hidden layer growing procedure to increase the network depth, $L=L+1$. The newly created layer $L$ is then trained in the low level learning phase using the current data batch $B_k$. Meanwhile, the network parameters of other hidden layers are frozen to preserve the old knowledge which prevents the catastrophic forgetting. The warning phase (\ref{equation:warning}) indicates the transition situation where more observations are required to signal a concept drift. Because of this reason, the current data batch is stored in the buffer $B_{warning}=B_k$ and is exploited to initialize a new hidden layer if the drift occurs in the next timestamp $k+1$. The stable condition yields to the adjustment of the current structure via the low level learning phase and the deletion of the data in the buffer.

\begin{figure*}[t!]
	\begin{centering}
	\includegraphics[scale=0.6]{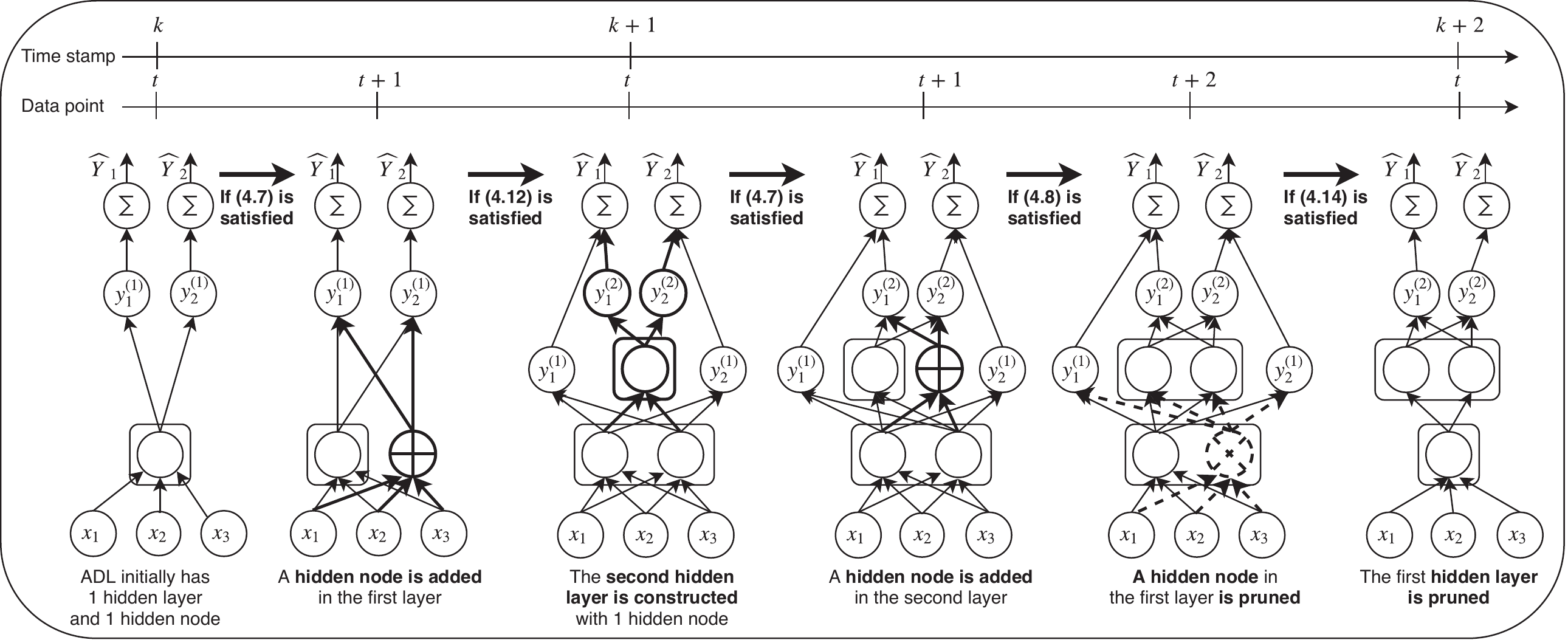}
	\par\end{centering}
	\caption{Example of an automated DNN construction used by the proposed ADL}
	\label{fig:evolution}
\end{figure*}

\paragraph{Hidden layer pruning.} ADL employs the hidden layer pruning mechanism to handle the redundancy across different hidden layers. This is achieved by analyzing the correlation of the output $y^{(l)}$ \cite{pensembleplus}. Based on the manifold learning concept, a redundant hidden layer embracing similar concept is expected not to inform an important representation of the given problem or at least very well covered by other hidden layers, because it does not open the manifold of learning problem to a unique representation. The MICI method is utilized to explore the correlation between two hidden layers it yields to the pruning condition, as follows:
\begin{equation} \label{equation:cpruning}
    \gamma(y^{(i)},y^{(j)})>\delta,\,\forall i,j=1,\dots,L, \,i \neq j
\end{equation}
$\delta$ is a user-defined threshold which is proportional to the maximum correlation index, where the lower the value the less pruning mechanism is executed. If (\ref{equation:cpruning}) is satisfied, the pruning process encompasses the hidden layer with the lowest $\beta$, i.e. $HLpruning \rightarrow \min_{l_p=i,j}\beta^{(l_p)}$. Note that $\beta$ is expected as an appropriate indicator of a hidden layer performance because it is dynamically adjusted using dynamic decreasing factor. Consequently, the direct connection from $l_p$-th hidden layer to the output $\hat{Y}$ is deleted yet that hidden layer still performs the forward-pass operation providing the representation $h^{(l_p)}$. This strategy also accelerates the model update because the pruned hidden layer is ignored in the learning procedure. It can be regarded as the dropout scenario in the realm of deep learning \cite{srivastava2014dropout}, yet ADL relies on the similarity analysis (\ref{equation:cpruning}) instead of the probabilistic approach. The illustration of incremental learning aspect of ADL is depicted in Figure \ref{fig:evolution}.

\subsection{The Solution of Catastrophic Forgetting}
Having a flexible structure embracing different-depth enables ADL to address the problem via two mechanisms elaborated in this section.

\paragraph{Dynamic voting weight adaptation.} Every voting weight $\beta^{(l)}$ is dynamically adjusted by a unique decreasing factor $p^{(l)} \in [0,1]$ which plays an important role while adapting to the concept drift. A high value of $p^{(l)}$ provides slow adaptation to the rapidly changing environment, yet it handles gradual or incremental drift very well. Conversely, a low value of $p^{(l)}$ gives frequent adaptation to sudden drift, yet it forfeits the stability while dealing with gradual drift where data samples embrace two distributions. This issue is handled by continuously adjusting $p^{(l)}$ to represent the performance of each hidden layer using a step size $\zeta$, as per in (\ref{equation:reward}). These are realized by setting $p^{(l)}$ to either $(p^{(l)} + \zeta)$ or $(p^{(l)} - \zeta)$ when the $l$-th hidden layer returns a correct prediction or incorrect one, respectively. This also considers the fact that the voting weight of a hidden layer embracing relevant representation should decrease slowly when making misclassification while that embracing irrelevant representation should increase slowly when returning the correct prediction.
\begin{eqnarray}\label{equation:reward}
    p^{(l)} = p^{(l)} \pm \zeta\\
    \beta^{(l)} = \min (\beta^{(l)}(1+p^{(l)}),1) \label{equation:rewarding}\\
    \beta^{(l)} = p^{(l)}.\beta^{(l)} \label{equation:penalizing}
\end{eqnarray}

The reward and penalty scenario are carried out by increasing and decreasing the voting weight based on the performance of its respective hidden layers, $y^{(l)}$. The reward is given when a hidden layer returns a correct prediction, as per in (\ref{equation:rewarding}). Conversely, a hidden layer is penalized if it makes an incorrect prediction, as per in (\ref{equation:penalizing}). The reward scenario is capable of handling the cyclic drift by reactivating the hidden layer embracing a small $\beta$. Unlike its predecessors in \cite{pENsemble,pensembleplus}, the aims of reward and penalty scenario carried out here are to augment the impact of a strong hidden layer by providing a high reward and a low penalty and to diminish a weak hidden layer by giving a small reward and a high penalty. Note that ADL possesses different-depth structure where every hidden layer has a direct connection to the output. As a result, the classification decision should consider the relevance of each hidden layer based on the prequential error. This approach aligns with the DDS as a method to increase the network depth because it guarantees ADL to embrace a different concept in each hidden layer.

\paragraph{Winning layer adaptation.} SGD method is employed to adjust the network parameters of the winning layer, i.e. $\theta^{(l_w)}$, using labelled data batch $B_k = (X_k,C_k) \in \Re^{T \times (n+m)}$ in a \textbf{single-pass manner}. It is derived using the cross-entropy loss minimization. However, instead of using the global error derivative, ADL exploits the local one which is backpropagated from the winning layer $y^{(l_w)}$. By this approach, each hidden layer is optimized based on a different objective which embraces a different concept. This enables ADL to improve its generalization power while reducing the risk of being suffered from catastrophic forgetting problem. Note that the parameter adjustment mechanism is executed under a dynamic network which consists of single hidden node in the beginning and can grow on demand.

\section{Empirical Evaluation}
This section outlines the empirical study of ADL in which it is compared against four algorithms.
\paragraph{Experimental setting.} DL is numerically validated using nine prominent data stream problems, i.e., Permuted MNIST \cite{kirkpatrick2016overcoming}, Weather \cite{DitzlerImbalanced}, KDDCup \cite{KDDCup}, SEA \cite{SEA}, hyperplane \cite{MOA}, SUSY, Hepmass \cite{Baldi2014SearchingFE}, RLCPS \cite{RLCPS}, and RFID localization \cite{DBLP:journals/corr/abs-1805-07715}. The first five characterize non-stationary properties, while the others feature prominent characteristics in examining the performance of data stream algorithms: big size, high input feature, etc. ADL is compared against fixed-structure DNN to observe kind of improvement produced by ADL while embracing the flexible different-depth structure. The value of $[\alpha_D,\alpha_W,\delta,\zeta]$ are set to $[0.0001,0.0005,0.05,0.001]$ in all problems. It is worth noting that $\alpha_D,\alpha_W$ determines the confidence level of Hoeffding bound $1-\alpha_{D,W}$. The selected values $0.0001,0.0005$ return very high confidence levels close to $100\%$. DNN network structure is initialized before the execution. ADL is compared to another deep stacked network embracing a flexible different-depth structure, that is DEVFNN (DFN) \cite{DEVFNN}. ADL is also compared against pEnsemble+ (pE+) \cite{pensembleplus} and pEnsemble (pE) \cite{pENsemble} aims to present the improvement over an evolving ensemble structure.
\begin{table}[!t]
\caption{Numerical results of consolidated algorithm}\label{table:numerical results}
\begin{centering}
\scalebox{0.8}{
\begin{tabular}{clrrrrr}
\toprule
 &  & Class. rate & ET & HL & HN & NoP \tabularnewline
\midrule 
S & ADL & $\textbf{78.26}\pm\textbf{2.8}$ & $2.5$K & $\textbf{2}\pm\textbf{0.6}$ & $614\pm{40}$ & $(17\pm{7})$K \tabularnewline
 
U & pE+ & $76.99\pm{4.6}$ & $35$K & $19\pm{6}$ & $9\pm{3}$ & $230\pm{80}$ \tabularnewline
 
S & pE & $74.44\pm{2.4}$ & $14$K & $3\pm{2}$ & $2\pm{1}$ & $36\pm{21}$ \tabularnewline
 
Y & DFN & $76.7\pm{3.2}$ & $5$K & $29\pm{12}$ & $24\pm{12}$ & $(72\pm{6.4})$K \tabularnewline
  
 & DNN & $51.19\pm{4.64}$ & $6$K & $23$ & $637$ & $5.7$K \tabularnewline

\midrule
H. & ADL & $\textbf{84.02}\pm\textbf{2.2}$ & $0.59$K & $\textbf{2}\pm\textbf{0.7}$ & $154\pm{10}$ & $(1.7\pm{1.3})$K \tabularnewline
 
M & pE+ & $82.3\pm{2.2}$ & $7.6$K & $2\pm{0.7}$ & $2\pm{0.7}$ & $24\pm{8}$ \tabularnewline
 
A & pE & $82.6\pm{1.9}$ & $12$K & $2\pm{0.7}$ & $2\pm{0.7}$ & $24\pm{8}$ \tabularnewline
 
S & DFN & $80.46\pm{6.87}$ & $1.5$K & $10\pm{5}$ & $8\pm{5}$ & $(15\pm{10})$K \tabularnewline
 
S & DNN & $50.03\pm{2.41}$ & $0.8$K & $8$ & $160$ & $4$K \tabularnewline
\midrule
R & ADL & $\textbf{99.99}\pm\textbf{0.03}$ & $1.6$K & $1$ & $58\pm{2}$ & $690\pm{20}$ \tabularnewline
 
L & pE+ & $99.8\pm{0.3}$ & $12.6$K & $7\pm{1}$ & $7\pm{1}$ & $84\pm{13}$ \tabularnewline
 
C & pE & $99.7\pm{0.3}$ & $60$K & $50\pm{15}$ & $50\pm{15}$ & $600\pm{190}$ \tabularnewline
 
P & DFN & $99.7\pm{0.3}$ & $0.8$K & $1$ & $1$ & $128$ \tabularnewline
 
S & DNN & $99.99\pm{0.03}$ & $0.54$K & $1$ & $58$ & $0.7$K \tabularnewline

\midrule
R & ADL & $\textbf{99.11}\pm\textbf{2}$ & $55.1$ & $1$ & $51\pm{10}$ & $420\pm{80}$ \tabularnewline
 
F & pE+ & $60.9\pm{7.6}$ & $0.7$K & $2\pm{0.8}$ & $1\pm{0.5}$ & $44\pm{14}$ \tabularnewline
 
I & pE & $60.4\pm{6.7}$ & $0.5$K & $2\pm{1}$ & $2\pm{0.7}$ & $43\pm{22}$ \tabularnewline
 
D & DFN & $93.5\pm{5.8}$ & $74.8$ & $10\pm{3}$ & $9\pm{3}$ & $(7.6\pm{5})$K \tabularnewline
 
 & DNN & $99.1\pm{2.7}$ & $33.02$ & $1$ & $51$ & $0.4$K \tabularnewline
\midrule
P. & ADL & $\textbf{81.62}\pm\textbf{11.5}$ & $26.3$ & $\textbf{1}$ & $22\pm{6}$ & $(18\pm{4.5})$K \tabularnewline
 
M & pE+ & NA & NA & NA & NA & NA \tabularnewline
 
NI & pE & NA & NA & NA & NA & NA \tabularnewline
 
S & DFN & NA & NA & NA & NA & NA \tabularnewline

T & DNN & $78.8\pm{12.7}$ & $18.5$ & $1$ & $20$ & $16$K \tabularnewline

\midrule
W & ADL & $74.48\pm{5.19}$ & $3.1$ & $1$ & $8\pm{1}$ & $93\pm{15}$ \tabularnewline
 
e & pE+ & $\textbf{78.8}\pm\textbf{4}$ & $29.42$ & $2\pm{0.2}$ & $1$ & $24\pm{2}$ \tabularnewline
 
a & pE & $78.4\pm{4.3}$ & $33.49$ & $2$ & $1$ & $24$ \tabularnewline
 
t & DFN & $78.6\pm{4.3}$ & $7.8$ & $3$ & $1$ & $318$ \tabularnewline
 
h. & DNN & $71.38\pm{8.74}$ & $1.98$ & $1$ & $8$ & $90$ \tabularnewline

\midrule
K & ADL & $\textbf{99.85}\pm\textbf{0.18}$ & $102.7$ & $1$ & $26\pm{1.2}$ & $1120\pm{50}$ \tabularnewline
 
D & pE+ & $96.7\pm{6}$ & $0.86$K & $1$ & $1$ & $12$ \tabularnewline
 
D & pE & $99.3\pm{0.4}$ & $5.4$K & $1$ & $1$ & $12$ \tabularnewline
 
C & DFN & $99.16\pm{0.5}$ & $0.21$K & $1$ & $1$ & $1900$ \tabularnewline
 
p. & DNN & $99.84\pm{0.19}$ & $62.46$ & $1$ & $26$ & $1000$ \tabularnewline
\midrule
S & ADL & $\textbf{92.82}\pm\textbf{5.79}$ & $18$ & $1$ & $22\pm{8}$ & $130\pm{50}$ \tabularnewline
 
E & pE+ & $92\pm{6}$ & $0.2$K & $5\pm{2}$ & $2\pm{1}$ & $60\pm{19}$ \tabularnewline
 
A & pE & $92\pm{5.7}$ & $0.18$K & $5\pm{2}$ & $2\pm{1}$ & $60\pm{19}$ \tabularnewline
 
 & DFN & $91.9\pm{5.3}$ & $39.1$ & $2$ & $1$ & $52$ \tabularnewline
 
 & DNN & $92.5\pm{6.49}$ & $11.28$ & $1$ & $22$ & $0.13$K \tabularnewline

\midrule
H & ADL & $\textbf{92.33}\pm\textbf{2.63}$ & $21.51$ & $1$ & $16\pm{1}$ & $110\pm{7}$ \tabularnewline
 
 y & pE+ & $87.6\pm{6.2}$ & $0.15$K & $5\pm{1}$ & $3\pm{0.5}$ & $55\pm{11}$ \tabularnewline
 
 p & pE & $91.8\pm{1.9}$ & $68.2$ & $4\pm{4}$ & $3\pm{2}$ & $23\pm{45}$ \tabularnewline
 
 e & DFN & $91.77\pm{1.6}$ & $0.3$K & $2$ & $1$ & $76$ \tabularnewline
 
 r. & DNN & $92\pm{3.28}$ & $13.39$ & $1$ & $16$ & $0.11$K \tabularnewline

\bottomrule
\end{tabular}}
\par\end{centering}
\centering{}ET: execution time, HL: hidden layers, HN: hidden nodes, NoP: number of parameters
\end{table}

The performance of all algorithms are assessed using five performance metrics: classification rate, execution time (ET), HL, HN, and the number of parameters (NoP). The prequential evaluation is conducted in a single-pass mode to simulate real data stream environments. The numerical results are averaged across all time stamps except the execution time. HL is the number of hidden-layer-to-output connection in ADL, the number of ensemble in pEnsemble and pEnsemble+, and the number of stacked building unit in DEVFNN. HN represents the total nodes in ADL and DNN, while in the remainder methods it signifies the total fuzzy rule. All experiments are executed in the same computational environment to assure fair comparisons under MATLAB environments with the Intel(R) Xeon(R) CPU E5-1650 @3.20 GHz processor and 16 GB RAM.

\paragraph{Numerical results.} From Table \ref{table:numerical results}, ADL delivers up to $68\%$ performance improvement over consolidated algorithms in terms of accuracy. This also demonstrates that the fully elastic network of ADL, where the hidden node and the hidden layer can be added or discarded on demand, can arrive at appropriate complexity for a specific problem and is comparable to those three evolving algorithms. ADL delivers the fastest execution time compared to those evolving algorithms in most cases. This result is understood because ADL is built upon MLP, while those algorithms are constructed by the multi-classifier concept possessing high computational and space complexity. This enables ADL to execute the high dimensional data, permuted MNIST problem, which results in $3\%$ improved accuracy over DNN, while the evolving algorithms are not scalable to deal with this problem. In terms of resolving the catastrophic forgetting, ADL delivers the most encouraging performance. The evidence can be seen from the numerical results of big datasets, SUSY and Hepmass, where ADL delivers the highest classification rate. These facts are reasonable since ADL characterizes different-depth structure supported by dynamic voting weight and winning layer adaptation which enables ADL to flexibly recall the previous knowledge or craft the new one.

\section{Conclusion}
This paper presents a novel self-organizing DNN, namely ADL. It possesses a flexible different-depth structure where the network structure can be automatically constructed from scratch with the absence of problem-specific user-defined parameters. The adaptation of network width is controlled by the estimation of bias and variance while the hidden layer can be deepened using the drift detection method. Possessing different-depth structure becomes the key characteristics of ADL to address catastrophic forgetting problem in the lifelong learning environment. It enables ADL to put more emphasis on the most relevant layer via dynamic voting scenario and winning layer adaptation. Our empirical evaluation has validated the effectiveness of ADL in dealing with non-stationary data streams under prequential test-then-train protocol. It also demonstrates the increase in performance over fixed structure DNN embracing the same network complexity. Future work inspired by this method should investigate the feasibility of ADL to handle unstructured data streams.

\section*{Acknowledgement}
This work is supported by A*STAR-NTU-SUTD AI Partnership Grant No. RGANS1902.

\bibliographystyle{unsrt}  
\bibliography{references}


\end{document}